\ifdefined\pdfinfoomitdate\pdfinfoomitdate=1\fi
\ifdefined\pdftrailerid\pdftrailerid{}\fi
\ifdefined\pdfsuppressptexinfo\pdfsuppressptexinfo=15\fi
\documentclass[letterpaper]{article}
\usepackage[preprint]{aaai2027}
\usepackage[hyphens]{url}
\usepackage{graphicx}
\usepackage{natbib}
\usepackage{caption}
\usepackage{booktabs}
\usepackage{amsmath}
\usepackage{amssymb}
\usepackage{array}
\usepackage{tabularx}
\usepackage{listings}
\usepackage{dblfloatfix}
\title{Success Is Not Self-Explanatory: Auditing Success Provenance in Agent Evaluation}
\author{
	Jingkun Luo,
	Da-Tian Peng\corresponding
}
\affiliations{
	Northwestern Polytechnical University\\
	luojingkun@mail.nwpu.edu.cn, pengdatian@nwpu.edu.cn
}

\begin{document}
	\maketitle
	
	\begin{abstract}
		A correct answer can conceal why an agent succeeded. Once agents change their information state during evaluation, correctness no longer distinguishes intended reasoning from answer acquisition. Outcome evidence and exposure detection do not establish whether success depended on an acquired target; we call this missing evaluation object success provenance. AcquaBench audits it through matched CLEAN, GOLD, and SHAM value substitution on four standardized surfaces with joint qid-clustered analysis. CLEAN retains benchmark-authorized information. GOLD makes the correct target available. SHAM preserves source structure and exposure opportunity but substitutes a matched incorrect value. GOLD minus CLEAN measures the total score response to correct-target availability; GOLD minus SHAM tests whether that response tracks target correctness beyond matched source exposure. In D0, GOLD exceeds SHAM by 19.1 to 25.9 percentage points, showing that success follows the correct value. In D2, GOLD still exceeds SHAM under distributed sufficiency while \textit{coloc} no longer transfers as a high-score marker, with AUROC 0.376 and 0.142. Behavioral dependence can thus persist beyond this probe's intended observation unit. In model comparison, a supported 5.0-point CLEAN score gap compresses to a raw GOLD difference of $-0.6$ points without establishing rank inversion. Agent benchmarks should report success together with whether the evaluated information state supported it. Code is available at \url{https://github.com/luojingkun22/acquabench}.
	\end{abstract}
	
	\begin{figure}[!t]
		\centering
		\includegraphics[width=0.95\columnwidth]{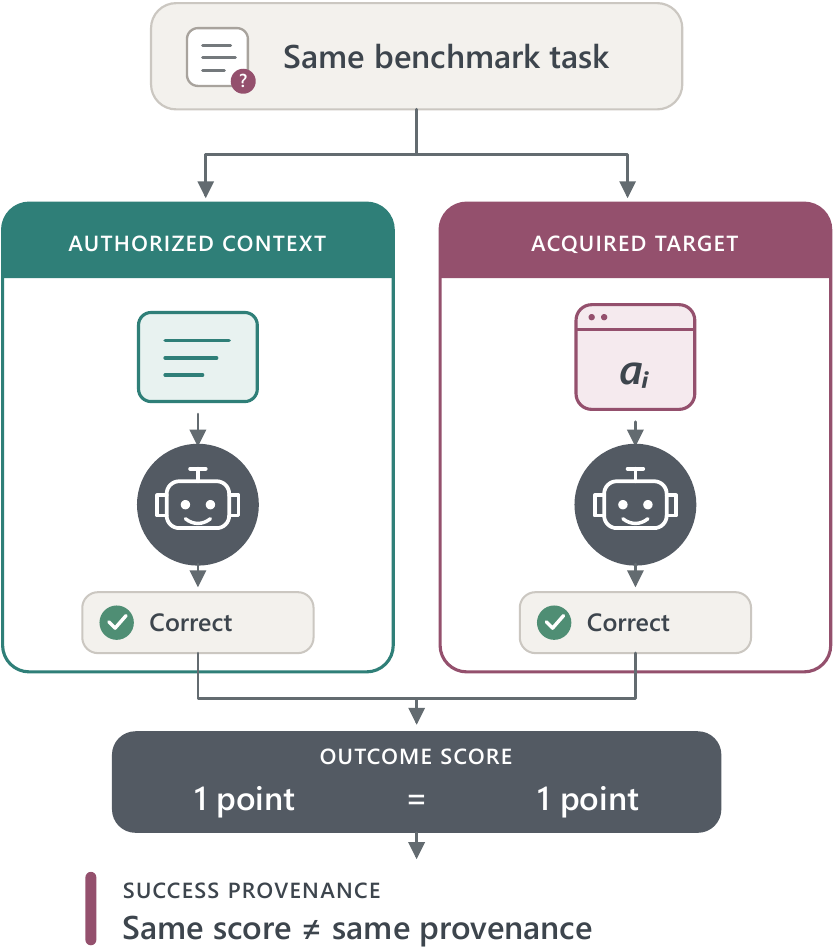}
		\caption{Why outcome-only scoring misses success provenance. A correct answer can follow either benchmark-authorized context or an acquired target value. Conventional scoring assigns both trajectories one point despite their different information provenance.}
		\label{fig:success-provenance}
	\end{figure}
	
	\section{Introduction}
	
	An agent can answer a benchmark question correctly for two fundamentally different reasons. It can solve the task from the information the benchmark intended to provide, or it can acquire answer-sufficient information during the evaluation itself. Conventional accuracy gives these trajectories the same score. This distinction becomes unavoidable as agent benchmarks permit systems to browse, call tools, retrieve external knowledge, and alter real or simulated environments \cite{liu2024agentbench,mialon2024gaia,yao2025taubench,wei2025browsecomp}. Outcome evidence establishes whether a claimed result occurred, and process-aware evaluation reveals how far an agent progressed \cite{ma2024agentboard,gao2026supportscores}. Neither asks whether the score depended on target information acquired inside the evaluated trajectory. Figure~\ref{fig:success-provenance} isolates the resulting gap: once the evaluated system can change its own information state, success is not self-explanatory.
	
	Contamination research asks whether test items entered model training and develops black-box tests for such exposure \cite{sainz-etal-2023-nlp,deng-etal-2024-investigating,oren2024proving,shi2024detecting,golchin2024timetravel}. Dynamic benchmarks refresh or transform evaluation items \cite{kiela-etal-2021-dynabench,white2025livebench,jain2025livecodebench}, while search-time work studies agents that retrieve public benchmark artifacts or answers during inference \cite{han2025searchtime,wang2026searchtime}. These lines establish exposure risk, but they do not test within-unit value dependence: holding the evaluated unit fixed, does the score change with the supplied target value? Naturally exposed qids can differ in difficulty, blocking a source can alter the surrounding retrieval path, and any additional source can change behavior even when its value is wrong. A suspicious trace is therefore evidence of exposure, not by itself evidence that success tracked the correct target value.
	
	We frame the missing evaluation object as \emph{success provenance}. A provenance audit asks whether the authorized information state already supported success and whether information acquired inside the evaluated trajectory changed that support. AcquaBench implements the audit through matched CLEAN, GOLD, and SHAM conditions within the same qid, model, reference target, action channel, and scorer. CLEAN retains the authorized information state. GOLD makes the correct target value available. SHAM preserves source structure and exposure opportunity while substituting a matched incorrect value. GOLD minus CLEAN measures the total score response to target availability. GOLD minus SHAM is a matched correct-versus-incorrect value-substitution contrast that tests whether the response tracks target correctness beyond source exposure alone. Retrieval, tool, memory, and subagent surfaces are controlled realizations of this intervention and are analyzed jointly, not treated as independent replications.
	
	The evidence moves from effect to boundary to consequence. Under direct single-source acquisition, GOLD exceeds SHAM by 19.1 to 25.9 percentage points across all three models, and the separation survives a style control that matches source form and wording. When answer sufficiency is distributed across two sources, GOLD retains gains of 11.8 and 14.6 points over SHAM while single-source \textit{coloc} no longer transfers as a high-score marker, with AUROC 0.376 and 0.142 against earned success. Behavioral dependence therefore persists after the acquisition route exceeds this probe's intended observation unit. The distinction changes model comparison: a supported 5.0-point CLEAN score advantage for Qwen2.5-32B over Qwen2.5-14B compresses to a raw GOLD difference of -0.6 points. The interval does not establish rank inversion, but the information condition obscures the supported CLEAN ordering.
	
	AcquaBench contributes success provenance as the missing evaluation object between exposure and outcome verification; a within-qid CLEAN/GOLD/SHAM value-substitution audit instantiated on four standardized surfaces with joint qid-clustered analysis; and evidence that target-value dependence can persist beyond a single-source probe's observation unit and distort a supported model comparison. It thus turns the remaining question---whether a score follows an evaluation-time supplied target---into a matched intervention.
	
	\section{Related Work}
	
	\subsection{Agent Evaluation and Evidence for Success}
	
	Interactive agent benchmarks now span web navigation \cite{deng2023mind2web,zhou2024webarena,drouin2024workarena}, tool execution \cite{qin2024toolllm,wang2024gta,yao2025taubench}, software and operating environments \cite{jimenez2024swebench,xie2024osworld,trivedi-etal-2024-appworld,rawles2025androidworld}, and general browsing or assistant tasks \cite{liu2024agentbench,mialon2024gaia,wei2025browsecomp}. These benchmarks establish whether an agent completes a task under a configured environment. They also make the agent's information state an active part of evaluation.
	
	AgentBoard adds progress and trajectory-level diagnostics beyond final success \cite{ma2024agentboard}, while AgentDojo tests whether untrusted tool content redirects behavior \cite{debenedetti2024agentdojo}. Evidence-supported evaluation asks whether outcome checks contain enough artifacts to verify a claimed environment change \cite{gao2026supportscores}. These are distinct audit layers. Task evaluation asks whether the agent succeeded. Outcome evidence asks whether the claimed result occurred. Security evaluation asks whether external content redirected behavior. Success provenance asks whether acquired target information supported the scored result. A benchmark can verify an outcome while leaving that dependence unresolved.
	
	\begin{figure*}[t]
		\centering
		\includegraphics[width=\textwidth]{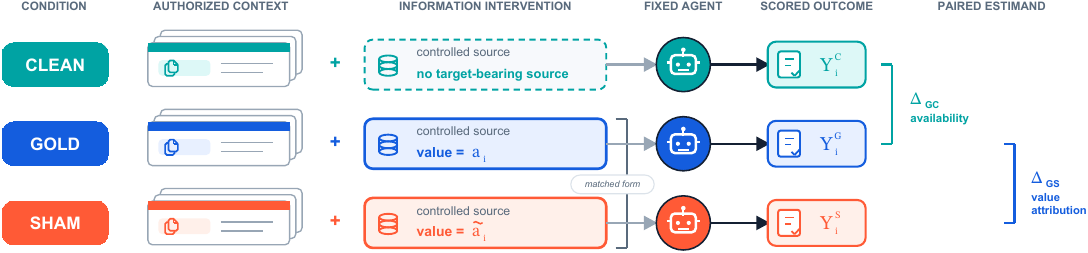}
		\caption{AcquaBench's matched value-substitution audit. CLEAN contains only benchmark-authorized information; GOLD adds a controlled source containing the correct target value; and SHAM preserves source structure and exposure opportunity while substituting a matched incorrect value. All conditions are matched within qid, model, reference target, action channel, and scorer. GOLD--CLEAN measures the total score response to correct-target availability; GOLD--SHAM tests whether that response tracks target correctness beyond matched source exposure.}
		\label{fig:matched-audit}
	\end{figure*}
	
	\subsection{Contamination Before and During Evaluation}
	
	Research on training data contamination asks whether benchmark examples entered model development data and how such exposure compromises held-out evaluation \cite{sainz-etal-2023-nlp,deng-etal-2024-investigating,oren2024proving,shi2024detecting,golchin2024timetravel,chen-etal-2025-benchmarking-large}. AcquaBench does not infer this hidden training history. CLEAN performance absorbs parametric knowledge and ordinary problem solving, while the audit isolates a controlled information change inside the evaluated trajectory.
	
	Dynamic data collection and controlled item transformation seek to keep evaluation informative as public benchmarks age \cite{kiela-etal-2021-dynabench,white2025livebench,jain2025livecodebench,chen2025dycodeeval,sun2025emperor}. Search-time contamination establishes the corresponding risk for agents that retrieve public benchmark questions and answers during inference \cite{han2025searchtime}, and subsequent work broadens detection across leakage types and measures performance inflation in deep research agents \cite{wang2026searchtime}. These studies detect suspicious sources, compare exposed subsets, or change source access. AcquaBench addresses a complementary identification problem: not whether an answer was exposed, but whether the score tracks the supplied target value. Within each qid, it preserves source structure and exposure opportunity between GOLD and SHAM while substituting that value. The contribution is not the discovery of contamination, but a matched audit of whether score changes follow the correct rather than a matched incorrect supplied value.
	
	\subsection{Counterfactual Attribution, Robustness, and Provenance}
	
	Contrast sets expose local decision boundaries through meaningful input perturbations that often change the reference label \cite{gardner-etal-2020-evaluating}. Retrieval robustness work introduces counterfactual noise to test whether models resist misleading evidence \cite{hong-etal-2024-gullible}. Counterfactual effect decomposition attributes sequential outcomes to agents, actions, or state variables \cite{triantafyllou2025counterfactual}. AcquaBench instead holds the question, target, model, action channel, and scorer fixed. SHAM is not a robustness benchmark. It is an attribution control that preserves source exposure while substituting the target value.
	
	Retrieval-augmented generation evaluates passage relevance and answer faithfulness \cite{es-etal-2024-ragas,saad-falcon-etal-2024-ares,niu-etal-2024-ragtruth}, while citation and provenance methods trace claims to supporting context \cite{gao-etal-2023-alce,sankararaman-etal-2024-provenance}. Agent-debugging methods instead use counterfactual replay or minimal repairs to locate failure-inducing steps \cite{shah2026causalagentreplay,bonagiri2026causalflow}. These lines connect outputs to supporting sources or responsible steps. AcquaBench changes the attribution unit to the evaluation-time information state. It tests whether behavioral dependence survives after answer sufficiency exceeds a single-source detector's observation unit and whether the resulting score change affects model comparison.
	
	\section{AcquaBench Design}
	
	\subsection{Auditing Success Provenance}
	
	AcquaBench turns the distinction above into an experimental variable. The same model can receive the same qid, use the same interface, and face the same reference target, yet succeed because an action made new target information available. The audit object is therefore a response together with the information path that preceded it. We call this success provenance. When acquisition outside the authorized information state changes the score, we refer to it as evaluation-time acquisition contamination. The term concerns information acquired inside the evaluated trajectory rather than hidden model-training history.
	
	Let $i$ index a fixed qid and action-channel pair. Let $Y_i^c$ denote binary success under information condition $c \in \{C,G,S\}$, and let $E_i^c$ indicate whether the controlled source was exposed before the final answer. Holding $i$ fixed makes the information condition the only intended difference within a triplet. The paired contrasts identify average score responses over the frozen population. They do not identify trajectory-specific causal effects. Trajectory-level labels provide operational case evidence, while population attribution rests on the matched GOLD minus SHAM contrast.
	
	\subsection{Paired Information-Condition Intervention}
	
	CLEAN exposes only the information authorized by the benchmark instance. GOLD preserves the qid, model, action interface, reference target, and scoring rule, then adds a matched source that makes the target value available. SHAM preserves the same interaction opportunity and source structure while substituting a matched incorrect value. The action channel remains fixed within every triplet, so retrieval is compared with retrieval, memory with memory, tool use with tool use, and subagent messaging with subagent messaging.
	
	The two paired contrasts serve different purposes. The quantity $\Delta_{GC}=\mathrm{E}[Y_i^G-Y_i^C]$ measures the total score response when the correct target becomes available. The quantity $\Delta_{GS}=\mathrm{E}[Y_i^G-Y_i^S]$ is a matched correct-versus-incorrect value-substitution effect: a positive value shows that the response tracks target correctness beyond matched source exposure. It does not separately identify benefit from the correct value and interference from the incorrect value. SHAM is not a no-information baseline. It preserves the opportunity to observe and use a matched source while substituting the value. Both contrasts are required because a gain over CLEAN alone cannot rule out generic interaction effects. Together, they operationalize the provenance distinction as a matched target-value attribution test. Figure~\ref{fig:matched-audit} summarizes the three conditions and estimands.
	
	\subsection{Acquisition Regimes and Premature Sufficiency}
	
	AcquaBench operationalizes direct acquisition through premature sufficiency. A source is prematurely sufficient when it exposes enough target information to resolve the question before the agent commits to its final answer, even though that information was unavailable in CLEAN. D0 instantiates this structure directly. A single GOLD source can be answer-sufficient, which makes direct acquisition behaviorally testable and potentially visible to a source-level detector.
	
	D1 tests a different boundary. It changes the answer form under a stricter surface constraint, so the measured outcome can depend on construction and scoring sensitivity. D1 is not designed to force a positive acquisition result. It asks whether the audit remains interpretable when the intervention also changes how an answer must be expressed.
	
	D2 removes single-source sufficiency without removing acquisition. It distributes a mapping and a relation across two sources. Neither source alone answers the question, but their composition can. D2 therefore separates the existence of behaviorally consequential acquisition from the visibility of that acquisition to a single-source detector. The three regimes are not pooled because each encodes a distinct information structure and supports a different inference.
	
	\begin{table*}[t]
		\centering
		\small
		\renewcommand{\arraystretch}{1.08}
		\setlength{\tabcolsep}{3pt}
		\begin{tabular}{@{}cp{0.25\textwidth}p{0.27\textwidth}p{0.36\textwidth}@{}}
			\toprule
			\multicolumn{1}{c}{Regime} & \multicolumn{1}{c}{Information structure} & \multicolumn{1}{c}{Frozen support} & \multicolumn{1}{c}{Role in audit} \\
			\midrule
			\textbf{D0} & Direct single-source sufficiency & 135/150 eligible qids $\times$ 4 channels; Llama-8B, Qwen-14B, Qwen-32B & Primary target-value dependence test and intended regime for single-source visibility. \\
			\textbf{D1} & Target-preserving surface-form variant & 38/135 D0-valid qids $\times$ 4 channels; Llama-8B, Qwen-32B & Construction and scorer-sensitivity boundary; not stable positive detector evidence. \\
			\textbf{D2} & Distributed two-source sufficiency & 36/50 parent qids $\times$ 4 channels; Llama-8B, Qwen-32B & Observation-unit boundary for the single-source detector; not an acquisition-null control. \\
			\bottomrule
		\end{tabular}
		\caption{Frozen AcquaBench regimes and eligible support. Every row uses matched CLEAN, GOLD, and SHAM trajectories across retrieval, tool use, memory access, and subagent messaging; regimes are analyzed separately.}
		\label{tab:regimes}
	\end{table*}
	
	\subsection{From Trajectories to Audit Evidence}
	
	At the trajectory level, an operationally attributable acquisition requires three events. CLEAN fails, GOLD succeeds, and the GOLD source is exposed before the final answer. This label identifies a concrete path for case analysis. It does not establish an individual causal effect or a population effect. Population attribution instead relies on the matched GOLD minus SHAM contrast.
	
	Among CLEAN failures, let $R_G=\mathrm{E}[Y_i^G\mid Y_i^C=0]$ and $R_S=\mathrm{E}[Y_i^S\mid Y_i^C=0]$. Conditional net rescue is $R_G-R_S$. It compares rescue rates over the same residual-failure population and is not a record-wise intersection of GOLD success with SHAM failure. We also report operationally cleaned accuracy by removing exposed GOLD successes that follow CLEAN failure from the raw GOLD score. This quantity is an audit view of the observed score, not a universally corrected leaderboard.
	
	The visibility analysis uses \textit{coloc}, a deliberately single-source question-answer score computed from the observed trajectory. For each distinct exposed source $s$, let $r_q(s)$ be the fraction of lowercase alphanumeric question tokens present in its title and text, and let $r_a(s)$ be the corresponding fraction of predicted-answer tokens. We define $\mathrm{coloc}=\max_s r_q(s)r_a(s)$. Sources are deduplicated by source identifier. The score is high only when one observed source jointly covers the qid and the model's eventual answer.
	
	Detector positives are GOLD trajectories that meet the operational attribution rule. The negative sets contain earned CLEAN successes---CLEAN trajectories that succeed without the controlled source---and successful exposed SHAM trajectories. This probe asks whether direct acquisition leaves the single-source structure predicted by premature sufficiency. D0 tests its intended regime, while D2 tests the boundary of that observation unit under composition. The analysis does not evaluate acquisition detectability in general.
	
	\section{Experimental Setup}
	
	The design above specifies the information intervention and the inferential role of each contrast. We now describe how the frozen benchmark populations, controlled artifacts, action surfaces, and estimators instantiate that design.
	
	\subsection{Benchmark Population and Frozen Evidence}
	
	We use the HotpotQA distractor development set, a multi-hop question answering benchmark built from Wikipedia passages \cite{yang-etal-2018-hotpotqa}. The frozen request set contains 150 question identifiers drawn with seed 13. Artifact eligibility is part of the experimental population rather than a post hoc filter. D0 contains 135 valid paired question identifiers. D1 is a strict eligible subset of 38 D0 identifiers. D2 contains 36 paired identifiers from a separately frozen 50-question parent sample.
	
	The eligibility restrictions are substantive limitations. D0 construction selects strongly toward shorter answers. D1 selects a narrower answer-type subset and is sensitive to answer form and scoring. We therefore condition all claims on the paired eligible populations. We do not pool D0, D1, and D2, and we do not combine supporting or development runs with canonical evidence.
	
	AcquaBench is a controlled identification benchmark, not an estimate of natural contamination prevalence. Its estimands are defined on frozen artifact-eligible populations within one benchmark family. The action channels are standardized surfaces for the same intervention rather than independent deployment environments. This deliberate scope preserves the interpretation of the matched contrasts while defining where the present claims apply.
	
	\subsection{Controlled Artifact Construction}
	
	Artifacts are generated and validated before agent inference, then frozen across models and action channels. D0 appends one short declarative claim to a selected base source. An accepted claim contains the target exactly once and directly resolves the question. GOLD fills the claim with the reference value. SHAM uses the same base source and claim template but substitutes a deterministic decoy selected to match answer type and approximate answer length. The paired artifacts therefore differ in target value rather than document shape.
	
	D1 replaces the canonical value with a verified alias or surface form while preserving its referent. D2 replaces direct sufficiency with two-source composition. One source links the question relation to an opaque key, and another maps that key to the target value. A D2 artifact is retained only when the maximum single-source \textit{coloc} for both GOLD and SHAM is at most 0.25. These checks encode the intended information structures before any outcome is observed.
	
	The style-controlled D0 validation targets construction artifacts directly. It removes the fixed title, short-document pattern, and repeated question wording. GOLD and SHAM use the same base document and declarative claim, with only the substituted value changing. Across the 135 paired identifiers, the base-document and claim-template match rates are 1.0, while the median paired differences in token count and question coverage are both zero. This run validates the D0 construction and is not treated as an independent dataset or model replication.
	
	\subsection{Models, Action Channels, and Scoring}
	
	The D0 build evaluates Llama-3.1-8B-Instruct and Qwen2.5-Instruct models at 14B and 32B parameters. The Llama family is documented in \cite{grattafiori2024llama3}, and the Qwen2.5 family in \cite{qwen2025qwen25}. D1 and D2 use Llama-3.1-8B and Qwen2.5-32B. The two Qwen sizes provide a prespecified matched within-family contrast. They do not establish a scaling law.
	
	Every paired qid is evaluated through retrieval, tool use, memory access, and subagent messaging. The four channels query the same BM25 evidence store and, for a fixed query, return the same top five ranked sources. They differ in action name and observation serialization, which makes them controlled interface surfaces rather than independent retrieval systems. Retrieval returns documents, tool use returns structured records, memory returns memory entries, and subagent messaging returns attributed reports. Main runs use deterministic decoding with temperature zero, at most eight action steps, and the two most recent observations in context. Sources are truncated to 1,800 characters and each step allows at most 256 generated tokens.
	
	Success is exact match or token F1 of at least 0.80. The same scorer and threshold are applied to CLEAN, GOLD, and SHAM. Table~\ref{tab:regimes} records the frozen population and intended inference for each regime.
	
	\subsection{Paired Statistical Analysis}
	
	All reported intervals use a paired qid-cluster percentile bootstrap with 5,000 replications and fixed analysis seed 13. A bootstrap draw resamples qids while retaining all associated channels and paired conditions. Model contrasts retain the prespecified common qid-channel support before resampling. This procedure preserves the dependence created by evaluating multiple action channels and information conditions for the same qid (full protocol and traceability: Supplementary Appendices A--B, E, and H).
	
	\begin{table*}[t]
		\centering
		\small
		\begin{tabular*}{\textwidth}{@{\extracolsep{\fill}}crrrrrr@{}}
			\toprule
			& \multicolumn{3}{c}{Observed success} & \multicolumn{3}{c}{Behavioral contrasts (pp)} \\
			\cmidrule(lr){2-4}\cmidrule(lr){5-7}
			\multicolumn{1}{c}{Model} & CLEAN & GOLD & SHAM & G-C & G-S & Net rescue \\
			\midrule
			Llama-8B & 286/540, 53.0\% & 383/540, 70.9\% & 243/540, 45.0\% & +18.0pp [12.4, 24.1] & +25.9pp [19.6, 32.4] & +31.9pp [22.5, 41.7] \\
			Qwen-14B & 362/540, 67.0\% & 453/540, 83.9\% & 324/540, 60.0\% & +16.9pp [11.9, 22.4] & +23.9pp [18.0, 30.2] & +38.8pp [27.6, 51.0] \\
			Qwen-32B & 389/540, 72.0\% & 450/540, 83.3\% & 347/540, 64.3\% & +11.3pp [6.7, 16.5] & +19.1pp [13.5, 24.8] & +37.7pp [24.5, 51.9] \\
			\bottomrule
		\end{tabular*}
		\caption{D0 results pooled over four standardized action surfaces. Success cells show $n/540$ and rate; contrast cells show percentage points with 95\% qid-cluster bootstrap intervals. G-C is GOLD minus CLEAN, G-S is GOLD minus SHAM, and Net rescue compares GOLD and SHAM on the same CLEAN-failed population.}
		\label{tab:d0-main}
	\end{table*}
	
	\section{Results}
	
	We report results in evidentiary order. D0 estimates target-value dependence and validates the high-\textit{coloc} signature in its intended regime; style control tests the strongest construction alternative. D1 isolates a measurement boundary, and D2 tests whether behavioral dependence persists when the information path exceeds a single-source observation unit. We then examine the consequence for a prespecified within-Qwen comparison.
	
	\subsection{D0 Success Tracks the Acquired Target Value}
	
	Table~\ref{tab:d0-main} establishes the value-attribution link. Each model contributes 540 paired qid-channel records from the same 135 qids. GOLD exceeds CLEAN for every model, showing that target availability changes the observed score. More importantly, GOLD also exceeds matched SHAM for every model. The score response therefore follows the correct target value rather than a source-shaped interaction alone.
	
	Across both model families, GOLD minus CLEAN ranges from 11.3 to 18.0 percentage points, GOLD minus SHAM from 19.1 to 25.9 points, and conditional net rescue from 31.9 to 38.8 points. All model-level intervals exclude zero, and all 12 prespecified model-surface GOLD--SHAM point estimates are descriptively positive (Supplementary Table S8). Confirmatory inference remains joint and qid-clustered, not four independent replications.
	
	The D0 behavior also has the structural signature predicted by direct acquisition. For Llama, \textit{coloc} reaches AUROC 0.929 [0.881, 0.966] against earned CLEAN success and 0.908 [0.848, 0.956] against successful exposed SHAM trajectories. The corresponding values are 0.925 [0.872, 0.967] and 0.908 [0.849, 0.957] for Qwen2.5-14B, then 0.968 [0.938, 0.990] and 0.958 [0.920, 0.986] for Qwen2.5-32B. Behavior and single-source visibility therefore align in the regime for which \textit{coloc} was defined. This serves as a regime-specific validity check for the probe rather than evidence for a general detector.
	
	\subsection{The D0 Attribution Survives Style Control}
	
	The strongest alternative explanation for D0 is that \textit{coloc} detects planted wording rather than an acquisition path. The style-controlled Llama retrieval validation directly targets that objection. Over 135 strictly paired question identifiers, CLEAN accuracy is 40.7\%, GOLD accuracy is 63.0\%, and SHAM accuracy is 38.5\%. The \textit{coloc} AUROC is 0.944 [0.895, 0.980] against earned success and 0.929 [0.870, 0.974] against successful SHAM trajectories.
	
	The behavioral separation and detector discrimination both persist after wording and source style are matched. This validation strengthens the interpretation of D0 without serving as an independent dataset or model replication.
	
	\subsection{D1 Establishes a Measurement Boundary}
	
	Table~\ref{tab:boundary-evidence} distinguishes the D1 measurement boundary from the D2 detector-scope boundary. D1 changes the surface form under which an answer must be produced and scored. Llama has GOLD minus CLEAN -2.0pp [-11.8, 7.9], whereas Qwen2.5-32B has -10.5pp [-20.4, -2.6]. The Qwen result establishes lower measured success under this construction, not that acquisition is generally harmful. Answer form and scorer sensitivity both change, so D1 defines a measurement boundary rather than a stable acquisition effect. Its detector estimates are descriptive, especially for Qwen2.5-32B, where only two trajectories meet the positive rule.
	
	\begin{figure}[!t]
		\centering
		\includegraphics[width=\columnwidth]{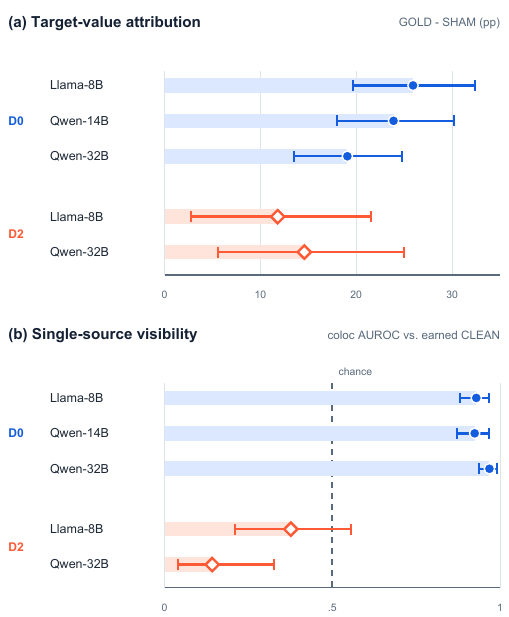}
		\caption{Behavioral value-substitution effects and single-source visibility across D0 and D2. (a) GOLD--SHAM contrasts in percentage points. (b) \textit{coloc} AUROC for distinguishing operationally attributable GOLD trajectories from earned CLEAN successes; the dashed line marks AUROC 0.5. Bars and whiskers show point estimates and 95\% qid-cluster percentile-bootstrap intervals. D0 combines a positive substitution effect with a high-\textit{coloc} signature. D2 retains a positive substitution effect while that score no longer transfers in its intended high-score direction under distributed sufficiency.}
		\label{fig:behavior-visibility}
	\end{figure}
	
	\begin{table*}[t]
		\centering
		\small
		\setlength{\tabcolsep}{2.6pt}
		\begin{tabular}{@{}ccccccc@{}}
			\toprule
			& & \multicolumn{3}{c}{Behavioral contrasts (pp)} & \multicolumn{2}{c}{Single-source \textit{coloc} AUROC} \\
			\cmidrule(lr){3-5}\cmidrule(lr){6-7}
			\multicolumn{1}{c}{\shortstack{Regime\\(qids)}} & \multicolumn{1}{c}{Model} & G-C & G-S & Net rescue & vs. earned & vs. SHAM \\
			\midrule
			D1 (38) & Llama-8B & \shortstack{-2.0pp \\ \mbox{[-11.8, 7.9]}} & \shortstack{+15.1pp \\ \mbox{[5.9, 24.3]}} & \shortstack{+12.1pp \\ \mbox{[1.4, 24.6]}} & \shortstack{0.457 \\ \mbox{[0.237, 0.768]} \\ $n=15/86$} & \shortstack{0.429 \\ \mbox{[0.208, 0.745]} \\ $n=15/59$} \\
			D1 (38) & Qwen-32B & \shortstack{-10.5pp \\ \mbox{[-20.4, -2.6]}} & \shortstack{-2.0pp \\ \mbox{[-7.2, 2.0]}} & \shortstack{+2.4pp \\ \mbox{[0.0, 9.7]}} & \shortstack{0.189 \\ \mbox{[0.045, 0.361]} \\ $n=2/111$} & \shortstack{0.171 \\ \mbox{[0.040, 0.340]} \\ $n=2/95$} \\
			\addlinespace[2pt]
			D2 (36) & Llama-8B & \shortstack{+13.9pp \\ \mbox{[4.9, 24.3]}} & \shortstack{+11.8pp \\ \mbox{[2.8, 21.5]}} & \shortstack{+8.3pp \\ \mbox{[-6.9, 26.9]}} & \shortstack{0.376 \\ \mbox{[0.210, 0.555]} \\ $n=25/84$} & \shortstack{0.360 \\ \mbox{[0.217, 0.510]} \\ $n=25/85$} \\
			D2 (36) & Qwen-32B & \shortstack{+4.2pp \\ \mbox{[-4.9, 13.2]}} & \shortstack{+14.6pp \\ \mbox{[5.6, 25.0]}} & \shortstack{+27.6pp \\ \mbox{[6.2, 57.1]}} & \shortstack{0.142 \\ \mbox{[0.040, 0.327]} \\ $n=12/115$} & \shortstack{0.139 \\ \mbox{[0.037, 0.324]} \\ $n=12/95$} \\
			\bottomrule
		\end{tabular}
		\caption{Boundary results. G-C is GOLD minus CLEAN; G-S is GOLD minus SHAM; Net rescue uses the shared CLEAN-failed population. Behavioral cells show percentage points [95\% CI]; AUROC cells show estimate [95\% CI] and $n=\mathrm{positive}/\mathrm{negative}$. D1 detector estimates are descriptive; D2 tests the detector's observation-unit boundary.}
		\label{tab:boundary-evidence}
	\end{table*}
	
	\subsection{D2 Separates Behavioral Dependence from Single-Source Visibility}
	
	D2 supplies the conceptual turn by distributing the necessary information across sources. Llama retains GOLD minus SHAM +11.8pp [2.8, 21.5]. Qwen2.5-32B retains +14.6pp [5.6, 25.0] and conditional net rescue +27.6pp [6.2, 57.1]. Target-value dependence therefore persists even though no individual source is sufficient by itself.
	
	The visibility boundary follows from that change in information structure. In D2, \textit{coloc} AUROC against earned success is 0.376 [0.210, 0.555] for Llama and 0.142 [0.040, 0.327] for Qwen2.5-32B. Against successful SHAM trajectories, it is 0.360 [0.217, 0.510] for Llama and 0.139 [0.037, 0.324] for Qwen2.5-32B. These below-0.5 estimates do not mean an absence of structure: operational positives receive lower single-source scores when sufficiency is distributed; the score distributions confirm this polarity (Supplementary Table S6). The high-\textit{coloc} rule therefore fails to transfer in its intended direction even as the behavioral effect persists. D2 is not an acquisition-null condition; it identifies the observation-unit boundary at which behavioral dependence and positively oriented single-source visibility separate. Figure~\ref{fig:behavior-visibility} aligns these properties across D0 and D2.
	
	\subsection{Direct and Compositional Paths Explain the Boundary}
	
	The aggregate results establish target-value dependence and its visibility boundary. Two prespecified trajectories make the structures concrete; their verified predictions, exposure order, EM/F1, and \textit{coloc} appear in Supplementary Tables S14--S15. They are explanatory, not additional statistical evidence.
	
	The D0 retrieval case asks ``The screenplay for Alain Resnais' second film is who?'' CLEAN predicts ``Alain Resnais, Chris Marker, and Ghislain Cloquet'' and fails. GOLD exposes one source that directly states Alain Robbe-Grillet, after which the model answers correctly. SHAM preserves the source structure but substitutes the value, and the prediction follows the substitution. D0 therefore exposes a direct path in which one source is answer-sufficient.
	
	The D2 subagent case asks ``Who is the film dedicated to that Paramount Classics and MTV Films co-purchased the rights to?'' CLEAN reports insufficient information. GOLD distributes the path: one source maps an opaque key to Sam Phillips, and a second states that the film is dedicated to that key. The model composes the mapping and relation to answer correctly. SHAM substitutes St. Vincent in the key mapping; the model answers ``Not specified in evidence'' and fails. No individual source answers the qid. The case explains why target-value dependence can persist while a high-\textit{coloc} single-source signature no longer transfers in its intended direction.
	
	\subsection{Exposed Scoring Compresses a Supported Model Gap}
	
	The evaluation consequence becomes visible when the same information intervention is applied to two sizes from the same model family. Table~\ref{tab:ranking-views} reports the matched differences. Under CLEAN, Qwen2.5-32B exceeds Qwen2.5-14B by +5.0pp [0.2, 10.2]. Under raw D0 GOLD evaluation, the gap becomes -0.6pp [-5.2, 4.1]. The point ordering reverses, although the interval does not establish rank inversion. The D0 inflation difference is -5.6pp [-10.9, -0.2]. The larger model shows less D0 inflation under correct-target availability, while the conditional net-rescue difference of -1.0pp [-14.4, 12.4] remains uncertain.
	
	The models and qids do not change across these views. Only the evaluation information condition changes. Raw GOLD evaluation nevertheless compresses the score gap supported under CLEAN and obscures its ordering. The operationally cleaned audit view returns the point ordering to +4.4pp, with 66.3\% for Qwen2.5-14B and 70.7\% for Qwen2.5-32B; it is not an inferentially established corrected gap. The supported conclusion is ranking-gap compression and distortion. The evidence does not establish a general scaling law, a significant rank inversion, or a universal corrected leaderboard.
	
	\begin{table}[!ht]
		\centering
		\small
		\begin{tabular*}{\columnwidth}{@{}l@{\extracolsep{\fill}}r@{\extracolsep{0pt}\hspace{0.6em}}l@{}}
			\toprule
			\multicolumn{1}{c}{Matched quantity} & \multicolumn{1}{c}{$\Delta$ (pp)} & \multicolumn{1}{c}{95\% CI} \\
			\midrule
			\multicolumn{3}{@{}l}{\textit{Observed score gap}} \\ 
			CLEAN score gap & +5.0pp & [0.2, 10.2] \\ 
			Raw GOLD score gap & -0.6pp & [-5.2, 4.1] \\ 
			Matched SHAM score gap & +4.3pp & [-0.7, 9.4] \\ 
			\addlinespace[3pt]
			\multicolumn{3}{@{}l}{\textit{Information-condition effect difference}} \\ 
			D0 inflation & -5.6pp & [-10.9, -0.2] \\ 
			GOLD--SHAM substitution & -4.8pp & [-10.0, 0.2] \\ 
			Conditional net rescue & -1.0pp & [-14.4, 12.4] \\ 
			\bottomrule
		\end{tabular*}
		\caption{Within-Qwen matched differences on common qid-channel support. $\Delta$ is Qwen2.5-32B minus Qwen2.5-14B, so positive values favor the larger model. The upper block compares observed score gaps under the three information conditions; the lower block compares information-condition effects. Intervals are 95\% qid-cluster bootstrap CIs.}
		\label{tab:ranking-views}
	\end{table}
	
	Together, the information intervention can differentially inflate model scores and compress a supported gap.
	
	\section{Conclusion}
	
	Correctness establishes the outcome, not why it became possible. AcquaBench audits this missing success provenance with within-qid CLEAN, GOLD, and SHAM conditions. CLEAN preserves the authorized information state; matched GOLD and SHAM preserve source structure but differ in whether the supplied value is correct. Their contrast tests whether success tracks target correctness beyond source exposure.
	
	Across frozen eligible populations, D0 shows that success follows the correct target value, and style control weakens alternatives based on wording and source form. D1 marks the measurement boundaries imposed by answer form and scoring. D2 supplies the central turn: behavioral target-value dependence persists when answer sufficiency is distributed across sources, even as the high-\textit{coloc} signature no longer transfers in its intended direction. Behavioral dependence and positively oriented single-source visibility are therefore distinct properties. The ranking analysis shows the evaluation consequence: exposed scoring can compress a model difference supported under CLEAN.
	
	These claims remain scoped to one benchmark family, frozen populations, and the probe's intended regime. Two Qwen sizes do not establish a scaling law, and operational cleaning is not a corrected leaderboard. Benchmark success is not self-explanatory: agent benchmarks should report whether the evaluated information state supported it.
	
	\bibliography{aaai2027}
	
	\setcounter{secnumdepth}{2}
\setcounter{table}{0}
\setcounter{figure}{0}
\renewcommand{\thetable}{S\arabic{table}}
\renewcommand{\thefigure}{S\arabic{figure}}
\newcommand{\CLEAN}{\textsc{Clean}}
\newcommand{\GOLD}{\textsc{Gold}}
\newcommand{\SHAM}{\textsc{Sham}}
\newcommand{\coloc}{\operatorname{coloc}}
\newcommand{\pp}{\ensuremath{\,\mathrm{pp}}}
\newcolumntype{Y}{>{\raggedright\arraybackslash}X}
\newcolumntype{R}{>{\raggedleft\arraybackslash}X}
\lstset{
	basicstyle=\scriptsize\ttfamily,
	columns=fullflexible,
	breaklines=true,
	breakatwhitespace=true,
	frame=single,
	aboveskip=4pt,
	belowskip=4pt,
	showstringspaces=false
}
	
	\section*{Supplementary Material}
	
	\section*{Scope and Reading Map}
	
	This supplement expands the construction, statistical protocol, complete frozen
	results, interpretation boundaries, reproduction details, model-comparison audit,
	mechanism cases, and provenance chain for AcquaBench. It does not introduce a new
	eligible population, model run, endpoint, or headline claim. All inferential results
	come from the same canonical trajectories and the same qid-cluster bootstrap used in
	the main paper. Descriptive per-surface and score-distribution summaries are reported
	only to expose heterogeneity and detector directionality; they are not treated as
	independent replications or new confirmatory endpoints.
	
	\appendix
	
	\section{Audit Construction and Validation}
	\label{app:construction}
	
	\subsection{Matched Evaluation Unit}
	
	The basic matched unit is a fixed question identifier (qid), model, reference target,
	action surface, and scorer. Let $i=(q,a)$ index a qid $q$ and action surface $a$.
	Within a triplet, the only intended manipulation is the controlled information
	condition $c\in\{C,G,S\}$. The model weights, question, reference answer, action name,
	source ranking procedure, maximum steps, decoding rule, and success scorer are fixed.
	
	\CLEAN{} exposes only the benchmark-authorized evidence. \GOLD{} replaces one or two
	selected evidence sources with frozen counterparts that make the correct target value
	available. \SHAM{} uses the same replacement positions, source shape, templates, and
	action surface, but substitutes a deterministic incorrect value. Thus \SHAM{} matches
	source structure and exposure opportunity; it does not assert identical realized
	exposure, attention, or subsequent action paths.
	
	\begin{table*}[t]
		\centering
		\small
		\begin{tabularx}{\textwidth}{@{}lrrrrYY@{}}
			\toprule
			Regime & Parent & Templates & Paired qids & qid-surface units & Models & Inferential role \\
			\midrule
			D0 direct & 150 & 135 & 135 & 540/model & Llama-3.1-8B; Qwen2.5-14B/32B & Direct single-source value substitution and intended high-$\coloc$ regime. \\
			D1 surface & 135 D0 & 42 alias-eligible & 38 & 152/model & Llama-3.1-8B; Qwen2.5-32B & Answer-form and scorer measurement boundary. \\
			D2 decomposed & 50 & 44 & 36 & 144/model & Llama-3.1-8B; Qwen2.5-32B & Distributed sufficiency and single-source observation-unit boundary. \\
			\bottomrule
		\end{tabularx}
		\caption{Frozen populations. ``Parent'' is the prespecified request set relevant to the
			construction stage; all reported behavioral estimates condition on the final paired
			artifact-eligible population. Four standardized surfaces are retained within each qid
			cluster rather than counted as independent samples.}
		\label{tab:populations}
	\end{table*}
	
	\subsection{Dataset Sampling and Authorized Evidence}
	
	The request sets were sampled from HotpotQA distractor development data after removing
	yes/no answers. A fixed pseudorandom shuffle with seed 13 selected the first 150 items
	for D0/D1 and a separately frozen 50-item parent sample for D2. \CLEAN{} retains all
	official distractor-context paragraphs and the official supporting-fact paragraphs.
	Retrieving an authorized source is therefore not itself contamination. The controlled
	intervention modifies evidence only through the frozen artifacts described below.
	
	Eligibility is part of the estimand rather than an outcome-dependent filter. D0 accepted
	135 of 150 qids; all 15 rejections were recorded as missing a verified direct template.
	The eligibility audit shows strong answer-length selection: accepted D0 answers contain
	approximately 2.46 words and 16 characters on average, versus 8.2 words and 49 characters
	among rejected items. D1 retains 38 of the 135 D0 qids for which both a gold surface form
	and the required matched decoy surface form pass deterministic and model-assisted checks.
	D2 retains 36 of 50 parent qids. Claims are therefore conditional on these frozen
	populations and do not estimate natural contamination prevalence.
	
	\subsection{D0: Direct Single-Source Sufficiency}
	
	For each qid, the constructor selects an authorized base source that does not already
	contain the normalized reference value. A short declarative template is generated and
	verified before agent inference. An accepted template must contain exactly one
	\texttt{<VALUE>} slot, directly resolve the factual question when filled, avoid the
	reference elsewhere, contain no question mark or prohibited meta-answer wording, and
	remain within the frozen length checks. \GOLD{} fills the slot with the reference;
	\SHAM{} fills it with a deterministic decoy.
	
	The decoy is selected from another frozen item. Candidates that normalize to the gold,
	contain the gold, or are contained by it are excluded. Remaining candidates are sorted
	by answer-type mismatch (numeric, named, or text), absolute normalized word-length
	difference, and qid. A SHA-256-derived index breaks ties deterministically. The gold and
	SHAM versions replace the same base source, retain its title, and differ only in the
	filled value.
	
	\subsection{D1: Surface-Form Boundary}
	
	D1 retains the D0 source and relation but fills the value slot with a directly readable
	surface form that preserves the referent while avoiding the normalized canonical phrase.
	Permitted transformations are verified date/number formatting changes and lexical
	shortenings whose tokens are present as full tokens, prefixes, or initials in the
	canonical value. A second verifier accepts only candidates that remain directly
	recoverable without factual inference. Both the reference and its matched decoy must
	have accepted forms. Examples include ``Peachtree St.'' for ``Peachtree Street'' and a
	short office title for a longer reference string.
	
	The outcome scorer still compares the model prediction with the canonical reference.
	Consequently, D1 jointly changes the evidence surface and the answer form presented to
	the model. Negative or unstable differences are evidence of this measurement boundary,
	not a general effect of harmful acquisition.
	
	\subsection{D2: Distributed Sufficiency}
	
	D2 decomposes a direct template into two sources. The relation source replaces the
	target with an opaque handle
	\[
	h_q=\texttt{Key-}\operatorname{prefix}_8(\operatorname{SHA256}(q)),
	\]
	while the mapping source states that $h_q$ denotes the target. \SHAM{} preserves the
	same handle, relation source, source titles, positions, and construction mode, but maps
	the handle to the deterministic decoy.
	
	The constructor first attempts to append each role to a selected base source. If either
	\GOLD{} or \SHAM{} would exceed the prespecified maximum single-source construction score
	of 0.25, both conditions switch to standalone role replacement. Of the 36 accepted qids,
	10 use append-to-base and 26 use standalone role replacement. For accepted artifacts,
	the \GOLD{} maximum single-source construction score ranges from 0 to 0.25 (mean 0.094),
	and the \SHAM{} score ranges from 0 to 0.25 (mean 0.069). The gate is applied before any
	behavioral outcome is observed.
	
	The construction score uses the supplied target value, whereas the runtime detector in
	Appendix~\ref{app:estimands} uses the model's predicted answer and only sources actually
	observed. The two calculations share a single-source observation unit but are not the
	same random variable. D2 therefore tests whether behavioral value dependence persists
	after answer sufficiency exceeds that observation unit; it does not test whether
	acquisition has no detectable structure in principle.
	
	\subsection{Four Standardized Action Surfaces}
	
	All surfaces query the same per-qid BM25 evidence store. For a fixed query they rank the
	same title-plus-text records and return the top five. They differ only in action name and
	observation serialization.
	
	\begin{table}[t]
		\centering
		\small
		\begin{tabularx}{\columnwidth}{@{}l l Y@{}}
			\toprule
			Surface & Action & Serialization \\
			\midrule
			Retrieval & \texttt{Search} & \texttt{[Document: title] text} \\
			Tool & \texttt{Lookup} & JSON records from \texttt{local\_evidence\_lookup} \\
			Memory & \texttt{ReadMemory} & XML-like \texttt{<memory>} entries \\
			Subagent & \texttt{AskSubagent} & Attributed local report lines \\
			\bottomrule
		\end{tabularx}
		\caption{Controlled interface realizations. These are not independent retrieval
			systems or deployment environments.}
		\label{tab:surfaces}
	\end{table}
	
	\subsection{Style-Controlled D0 Validation}
	
	The style validation addresses the strongest artifact alternative: a detector might
	recognize a fixed planted title, an unusually short source, or repeated question wording
	rather than the acquisition path. It uses Llama-3.1-8B on retrieval, with 135 strictly
	paired qids. \GOLD{} and \SHAM{} share the same base document and declarative template;
	only the filled value changes. Base-document and claim-template match rates are both
	1.0. The median absolute paired difference is zero for planted token count and question
	coverage. This is construction validation on the same experimental lineage, not an
	independent benchmark or model replication.
	
	\section{Estimands and Statistical Protocol}
	\label{app:estimands}
	
	\subsection{Outcome and Pairing}
	
	For each qid-surface unit $i$ and information condition $c$, binary success is
	\[
	Y_i^c=\mathbb{1}\{\operatorname{EM}=1\ \lor\ \operatorname{tokenF1}\geq0.80\}.
	\]
	Exact match and token F1 use lowercase normalization, punctuation removal, removal of
	articles \{a, an, the\}, whitespace normalization, and multiset token overlap for F1.
	The same reference and scorer are used in all three conditions.
	
	\subsection{Paired Information-Condition Contrasts}
	
	Over the frozen population $\mathcal{I}$,
	\begin{align}
		\Delta_{GC} &= |\mathcal{I}|^{-1}\sum_{i\in\mathcal{I}}(Y_i^G-Y_i^C),\\
		\Delta_{GS} &= |\mathcal{I}|^{-1}\sum_{i\in\mathcal{I}}(Y_i^G-Y_i^S).
	\end{align}
	$\Delta_{GC}$ is the total score response when the correct target becomes available
	relative to authorized evidence. $\Delta_{GS}$ is the matched correct-versus-incorrect
	value-substitution effect. A positive $\Delta_{GS}$ shows that the score responds to
	target correctness beyond matched source structure and exposure opportunity.
	
	It does not separately identify help from the correct value and interference from the
	incorrect value. Algebraically,
	\[
	\Delta_{GS}=\Delta_{GC}-\Delta_{SC},\qquad
	\Delta_{SC}=\mathbb{E}[Y^S-Y^C].
	\]
	Thus a large $\Delta_{GS}$ can combine a \GOLD{} increase and a \SHAM{} decrease. The
	three-condition design identifies value sensitivity at the population level, not a
	trajectory-specific causal effect or a pure decomposition of help and harm.
	
	\subsection{Conditional Net Rescue}
	
	Let $\mathcal{F}=\{i\in\mathcal{I}:Y_i^C=0\}$ be the matched residual population that
	fails under \CLEAN{}. Define
	\[
	R_G=|\mathcal{F}|^{-1}\sum_{i\in\mathcal{F}}Y_i^G,\quad
	R_S=|\mathcal{F}|^{-1}\sum_{i\in\mathcal{F}}Y_i^S,
	\]
	and conditional net rescue as $R_G-R_S$. This is a difference between two rescue rates
	with the same \CLEAN-failed denominator. It is not the record-wise fraction satisfying
	$Y_i^G=1$ and $Y_i^S=0$.
	
	\subsection{Operational Trajectory Labels and Audit Cleaning}
	
	An ``earned'' negative is a successful \CLEAN{} trajectory. An operationally
	``attributable acquired'' trajectory satisfies three record-level conditions: its
	matched \CLEAN{} run fails, its \GOLD{} run succeeds, and at least one controlled source
	is observed before the final answer. Other labels include ambiguous (both \CLEAN{} and
	\GOLD{} succeed with exposure), exposed-failed, and not-exposed. These labels select
	auditable cases; the aggregate matched contrast supplies the population attribution.
	
	The operationally cleaned \GOLD{} view subtracts the count of operationally attributable
	\GOLD{} successes from the raw \GOLD{} numerator. It is a descriptive audit view, not an
	inferentially established corrected capability score or universal leaderboard.
	
	\subsection{Single-Source Visibility Score and Detector Sets}
	
	For each distinct observed source $s$, let $T_q$ be the normalized content-token set of
	the question (stopwords and tokens of length at most two removed), let $T_a$ be the
	normalized alphanumeric tokens of the predicted answer, and let $T_s$ be the tokens in
	the source title and text. Define
	\[
	r_q(s)=\frac{|T_q\cap T_s|}{|T_q|},\qquad
	r_a(s)=\frac{\sum_{t\in T_a}\mathbb{1}[t\in T_s]}{|T_a|},
	\]
	and
	\[
	\coloc=\max_{s\in\mathcal{S}_{\mathrm{observed}}}r_q(s)r_a(s).
	\]
	Sources are deduplicated by source identifier. The score is intentionally oriented so
	that high values indicate one observed source jointly covering the question and the
	model's eventual answer. Because $T_a$ is formed from the predicted answer, the score
	can also vary with answer surface form and verbosity; it is not a content-invariant
	measure of acquisition.
	
	Detector positives are operationally attributable \GOLD{} trajectories. Two negative
	sets are reported separately: earned \CLEAN{} successes and successful, exposed \SHAM{}
	trajectories. AUROC uses average ranks for ties. No threshold is tuned and D0, D1, and
	D2 scores are never pooled.
	
	\subsection{Bootstrap and Model Common Support}
	
	Every headline interval is a two-sided percentile 95\% interval from 5,000 bootstrap
	replications with analysis seed 13. A draw samples qids with replacement and retains all
	associated surfaces and matched conditions, preserving within-qid dependence. Detector
	intervals use the same qid clustering. The Qwen2.5-32B minus Qwen2.5-14B comparison first
	restricts both models to the identical 135-qid by four-surface support, then resamples
	those qids jointly. No channel is selected after viewing outcomes.
	
	\section{Complete Frozen Results}
	\label{app:results}
	
	\subsection{D0 Primary Results}
	
	\begin{table*}[t]
		\centering
		\scriptsize
		\begin{tabular}{@{}lccc ccc@{}}
			\toprule
			Model & \CLEAN{} & \GOLD{} & \SHAM{} & $\Delta_{GC}$ [95\% CI] & $\Delta_{GS}$ [95\% CI] & Net rescue [95\% CI] \\
			\midrule
			Llama-3.1-8B & 286/540 (53.0\%) & 383/540 (70.9\%) & 243/540 (45.0\%) & +18.0 [12.4, 24.1] & +25.9 [19.6, 32.4] & +31.9 [22.5, 41.7] \\
			Qwen2.5-14B & 362/540 (67.0\%) & 453/540 (83.9\%) & 324/540 (60.0\%) & +16.9 [11.9, 22.4] & +23.9 [18.0, 30.2] & +38.8 [27.6, 51.0] \\
			Qwen2.5-32B & 389/540 (72.0\%) & 450/540 (83.3\%) & 347/540 (64.3\%) & +11.3 [6.7, 16.5] & +19.1 [13.5, 24.8] & +37.7 [24.5, 51.9] \\
			\bottomrule
		\end{tabular}
		\caption{Complete D0 behavioral results. Effects are percentage points. All intervals
			are qid-cluster percentile-bootstrap intervals from the frozen report.}
		\label{tab:d0-complete}
	\end{table*}
	
	\subsection{D1 and D2 Boundary Results}
	
	\begin{table*}[t]
		\centering
		\scriptsize
		\begin{tabular}{@{}llccc ccc@{}}
			\toprule
			Regime & Model & \CLEAN{} & \GOLD{} & \SHAM{} & $\Delta_{GC}$ [95\% CI] & $\Delta_{GS}$ [95\% CI] & Net rescue [95\% CI] \\
			\midrule
			D1 & Llama-3.1-8B & 86/152 (56.6\%) & 83/152 (54.6\%) & 60/152 (39.5\%) & -2.0 [-11.8, 7.9] & +15.1 [5.9, 24.3] & +12.1 [1.4, 24.6] \\
			D1 & Qwen2.5-32B & 111/152 (73.0\%) & 95/152 (62.5\%) & 98/152 (64.5\%) & -10.5 [-20.4, -2.6] & -2.0 [-7.2, 2.0] & +2.4 [0.0, 9.7] \\
			\addlinespace
			D2 & Llama-3.1-8B & 84/144 (58.3\%) & 104/144 (72.2\%) & 87/144 (60.4\%) & +13.9 [4.9, 24.3] & +11.8 [2.8, 21.5] & +8.3 [-6.9, 26.9] \\
			D2 & Qwen2.5-32B & 115/144 (79.9\%) & 121/144 (84.0\%) & 100/144 (69.4\%) & +4.2 [-4.9, 13.2] & +14.6 [5.6, 25.0] & +27.6 [6.2, 57.1] \\
			\bottomrule
		\end{tabular}
		\caption{Complete D1 and D2 behavioral results. Effects are percentage points. D1 is
			a measurement boundary; D2 preserves a positive value-substitution effect in both
			models while changing the information structure.}
		\label{tab:boundary-complete}
	\end{table*}
	
	\subsection{Detector Results and Directionality}
	
	\begin{table*}[t]
		\centering
		\scriptsize
		\begin{tabular}{@{}llrrc rc@{}}
			\toprule
			Regime & Model & $n_+$ & Earned $n_-$ & AUROC vs earned [95\% CI] & SHAM $n_-$ & AUROC vs SHAM [95\% CI] \\
			\midrule
			D0 & Llama-3.1-8B & 112 & 286 & 0.929 [0.881, 0.966] & 234 & 0.908 [0.848, 0.956] \\
			D0 & Qwen2.5-14B & 95 & 362 & 0.925 [0.872, 0.967] & 314 & 0.908 [0.849, 0.957] \\
			D0 & Qwen2.5-32B & 68 & 389 & 0.968 [0.938, 0.990] & 336 & 0.958 [0.920, 0.986] \\
			\addlinespace
			D1 & Llama-3.1-8B & 15 & 86 & 0.457 [0.237, 0.768] & 59 & 0.429 [0.208, 0.745] \\
			D1 & Qwen2.5-32B & 2 & 111 & 0.189 [0.045, 0.361] & 95 & 0.171 [0.040, 0.340] \\
			\addlinespace
			D2 & Llama-3.1-8B & 25 & 84 & 0.376 [0.210, 0.555] & 85 & 0.360 [0.217, 0.510] \\
			D2 & Qwen2.5-32B & 12 & 115 & 0.142 [0.040, 0.327] & 95 & 0.139 [0.037, 0.324] \\
			\bottomrule
		\end{tabular}
		\caption{Single-source $\coloc$ results. $n_+$ denotes operationally attributable
			\GOLD{} trajectories. The score is prespecified in the high-score direction. D1 is
			descriptive, particularly for Qwen2.5-32B with only two positives.}
		\label{tab:detector-complete}
	\end{table*}
	
	In D2, AUROC below 0.5 does not mean absence of any ordering signal. It means the
	prespecified rule ``higher single-source $\coloc$ implies direct acquisition'' does not
	transfer monotonically. The descriptive distributions in Table~\ref{tab:score-dist}
	make the direction explicit: operational positives receive lower scores than earned
	or successful-SHAM negatives, especially for Qwen2.5-32B. Reversing the score after
	observing this result would answer a different, post-hoc classification question and
	is not reported as a new detector.
	
	\begin{table*}[t]
		\centering
		\scriptsize
		\begin{tabular}{@{}llccc@{}}
			\toprule
			Regime & Model & Attributable \GOLD{} $n$: median [IQR] & Earned \CLEAN{} $n$: median [IQR] & Exposed successful \SHAM{} $n$: median [IQR] \\
			\midrule
			D0 & Llama-3.1-8B & 112: 0.838 [0.778, 0.909] & 286: 0.600 [0.467, 0.667] & 234: 0.613 [0.467, 0.688] \\
			D0 & Qwen2.5-14B & 95: 0.842 [0.757, 0.905] & 362: 0.565 [0.417, 0.643] & 314: 0.587 [0.455, 0.667] \\
			D0 & Qwen2.5-32B & 68: 0.882 [0.800, 0.915] & 389: 0.583 [0.455, 0.667] & 336: 0.600 [0.453, 0.667] \\
			\addlinespace
			D2 & Llama-3.1-8B & 25: 0.583 [0.438, 0.609] & 84: 0.611 [0.445, 0.667] & 85: 0.609 [0.500, 0.667] \\
			D2 & Qwen2.5-32B & 12: 0.318 [0.143, 0.448] & 115: 0.600 [0.477, 0.667] & 95: 0.600 [0.455, 0.667] \\
			\bottomrule
		\end{tabular}
		\caption{Read-only descriptive $\coloc$ distributions from the exact hashed canonical
			trajectory files. These summaries explain AUROC polarity and are not new inferential
			endpoints. D1 is omitted because it is not treated as positive detector evidence.}
		\label{tab:score-dist}
	\end{table*}
	
	\subsection{Style-Control Validation}
	
	\begin{center}\captionsetup{type=table}
		\centering
		\small
		\begin{tabular}{@{}lr@{}}
			\toprule
			Quantity & Frozen value \\
			\midrule
			Paired qids & 135 \\
			\CLEAN{} success & 55/135 (40.7\%) \\
			Style-\GOLD{} success & 85/135 (63.0\%) \\
			Style-\SHAM{} success & 52/135 (38.5\%) \\
			\GOLD{} rescue among 80 \CLEAN{} failures & 43.8\% \\
			\SHAM{} rescue among 80 \CLEAN{} failures & 13.8\% \\
			Net rescue & +30.0\pp \\
			Attributable positives & 35 \\
			$\coloc$ AUROC vs 55 earned & 0.944 [0.895, 0.980] \\
			$\coloc$ AUROC vs 52 SHAM successes & 0.929 [0.870, 0.974] \\
			String-match AUROC vs earned & 0.509 [0.500, 0.527] \\
			\bottomrule
		\end{tabular}
		\caption{Style-controlled D0 retrieval validation. The matched construction preserves
			the D0 pattern while naive string match remains near chance.}
		\label{tab:style-validation}
	\end{center}
	
	\subsection{Descriptive Results by Action Surface}
	
	Tables~\ref{tab:d0-surface} and~\ref{tab:boundary-surface} report every prespecified
	surface, not a favorable subset. They use the same paired qids as the pooled analyses.
	No surface-level confidence interval is promoted as a headline endpoint; confirmatory
	inference remains the joint qid-clustered analysis.
	
	\begin{table*}[t]
		\centering
		\scriptsize
		\begin{tabular}{@{}llrrrr@{}}
			\toprule
			Model & Surface & $\Delta_{GC}$ & $\Delta_{GS}$ & $\Delta_{SC}$ & Net rescue \\
			\midrule
			Llama-3.1-8B & Memory & +20.7 & +28.9 & -8.1 & +38.5 \\
			& Retrieval & +17.8 & +22.2 & -4.4 & +26.2 \\
			& Subagent & +18.5 & +23.0 & -4.4 & +30.9 \\
			& Tool & +14.8 & +29.6 & -14.8 & +32.1 \\
			\addlinespace
			Qwen2.5-14B & Memory & +20.7 & +27.4 & -6.7 & +40.8 \\
			& Retrieval & +15.6 & +20.0 & -4.4 & +37.2 \\
			& Subagent & +12.6 & +20.0 & -7.4 & +33.3 \\
			& Tool & +18.5 & +28.1 & -9.6 & +43.2 \\
			\addlinespace
			Qwen2.5-32B & Memory & +17.0 & +22.2 & -5.2 & +41.9 \\
			& Retrieval & +6.7 & +16.3 & -9.6 & +31.4 \\
			& Subagent & +10.4 & +14.8 & -4.4 & +36.8 \\
			& Tool & +11.1 & +23.0 & -11.9 & +40.0 \\
			\bottomrule
		\end{tabular}
		\caption{D0 descriptive effects by standardized surface, in percentage points. All
			12 $\Delta_{GS}$ point estimates are positive. Surfaces share the same evidence store
			and do not constitute independent replications.}
		\label{tab:d0-surface}
	\end{table*}
	
	\begin{table*}[t]
		\centering
		\scriptsize
		\begin{tabular}{@{}lllrrrr@{}}
			\toprule
			Regime & Model & Surface & $\Delta_{GC}$ & $\Delta_{GS}$ & $\Delta_{SC}$ & Net rescue \\
			\midrule
			D1 & Llama-3.1-8B & Memory & +2.6 & +13.2 & -10.5 & +11.1 \\
			& & Retrieval & -2.6 & +13.2 & -15.8 & +6.3 \\
			& & Subagent & 0.0 & +7.9 & -7.9 & +16.7 \\
			& & Tool & -7.9 & +26.3 & -34.2 & +14.3 \\
			D1 & Qwen2.5-32B & Memory & -10.5 & -2.6 & -7.9 & 0.0 \\
			& & Retrieval & -7.9 & -2.6 & -5.3 & 0.0 \\
			& & Subagent & -10.5 & +2.6 & -13.2 & +9.1 \\
			& & Tool & -13.2 & -5.3 & -7.9 & 0.0 \\
			\addlinespace
			D2 & Llama-3.1-8B & Memory & +16.7 & +13.9 & +2.8 & +11.8 \\
			& & Retrieval & +19.4 & +5.6 & +13.9 & +6.7 \\
			& & Subagent & +8.3 & +13.9 & -5.6 & +6.7 \\
			& & Tool & +11.1 & +13.9 & -2.8 & +7.7 \\
			D2 & Qwen2.5-32B & Memory & +5.6 & +13.9 & -8.3 & +25.0 \\
			& & Retrieval & -2.8 & +11.1 & -13.9 & +33.3 \\
			& & Subagent & +5.6 & +13.9 & -8.3 & +37.5 \\
			& & Tool & +8.3 & +19.4 & -11.1 & +14.3 \\
			\bottomrule
		\end{tabular}
		\caption{D1 and D2 descriptive effects by standardized surface, in percentage points.
			D1 exposes answer-form instability. All eight D2 $\Delta_{GS}$ point estimates are
			positive, while $\Delta_{GC}$ is not uniformly positive.}
		\label{tab:boundary-surface}
	\end{table*}
	
	\section{Interpretation and Claim Boundaries}
	\label{app:boundaries}
	
	\begin{table*}[t]
		\centering
		\small
		\begin{tabularx}{\textwidth}{@{}Y Y Y Y@{}}
			\toprule
			Evidence & Supported statement & Stronger unsupported statement & Scope condition \\
			\midrule
			$\Delta_{GS}>0$ & Scores respond differently to a correct versus matched incorrect supplied value. & Pure correct-value help; separate help/harm decomposition; individual causal proof. & Frozen matched population; SHAM matches structure and opportunity, not realized paths. \\
			D0 high-$\coloc$ AUROC & High single-source colocation separates operational positives in the direct regime. & A general acquisition detector. & Prespecified high-score direction and D0 observation unit. \\
			D2 positive $\Delta_{GS}$ with AUROC $<0.5$ & Behavioral value dependence persists while the D0 high-score rule fails to transfer monotonically. & Acquisition is invisible, undetectable, or signal-free in principle. & Two-source construction with a single-source gate. \\
			D1 results & Surface form and canonical-answer scoring can change measured success. & Acquisition is generally harmful. & 38-qid alias-eligible subset; exact-match/F1 scorer. \\
			Style validation & The D0 result survives matched base document, template, length, and coverage controls. & Independent dataset/model replication. & Llama retrieval on the same lineage. \\
			Within-Qwen audit & Correct-target availability differentially inflates the two sizes and compresses a supported \CLEAN{} gap. & Significant rank inversion or a scaling law. & Two Qwen sizes on common D0 support. \\
			Operational cleaning & Removing operationally attributable cases yields an auditable point-estimate view. & A causally corrected leaderboard or true capability ranking. & Record-level operational labels are not individual causal proofs. \\
			\bottomrule
		\end{tabularx}
		\caption{Claim-boundary matrix. The intended reading preserves the strongest supported
			claim without converting a scoped audit into a universal mechanism statement.}
		\label{tab:claim-boundaries}
	\end{table*}
	
	The central claim is not weakened by these boundaries. Outcome-only scoring cannot tell
	whether the authorized information state supported a success. The matched intervention
	shows, within a frozen evaluation unit, whether the score follows the correctness of a
	supplied target value. D2 then shows that this behavioral dependence and a particular
	high-single-source-visibility rule are distinct properties.
	
	\section{Reproducibility and Provenance}
	\label{app:reproduction}
	
	\subsection{Models and Runtime Configuration}
	
	The behavioral runs use the public Meta-Llama-3.1-8B-Instruct,
	Qwen2.5-14B-Instruct, and Qwen2.5-32B-Instruct checkpoints. Archived configurations
	load local, unquantized copies
	with their own tokenizers, \texttt{trust\_remote\_code=True}, bfloat16 weights, one-way
	tensor/pipeline/data parallelism, and an 8,192-token model window. The preserved,
	unchanged experiment environment records Python 3.12.3, PyTorch 2.11.0+cu130,
	vLLM 0.23.0, Transformers 5.5.4, and \texttt{rank-bm25} 0.2.2. Artifact template
	generation uses the same Llama checkpoint; sentence-similarity diagnostics use
	\texttt{all-MiniLM-L6-v2}, while the reported $\coloc$ endpoint is lexical. Runs used
	one NVIDIA RTX PRO 6000 GPU (96 GB), a single CUDA device with
	vLLM GPU-memory utilization set to 0.9, no quantization, and no model sharding. The
	release environment manifest retains the full package, driver, CUDA, and system details.
	
	Main behavioral runs use temperature 0, at most eight action steps, at most 256 generated
	tokens per step, the two most recent observations in rolling context, top-five BM25
	retrieval, and at most 1,800 displayed characters per source. If a source exceeds this
	limit, 60\% of the budget is taken from its beginning and the remainder from its end.
	The style-control run uses the same data/model family and seed with a six-step limit.
	
	\subsection{Agent Prompt and Generation Protocol}
	
	The action placeholder below is replaced by \texttt{Search}, \texttt{Lookup},
	\texttt{ReadMemory}, or \texttt{AskSubagent}. The model tokenizer's chat template is
	applied to the system and user messages with a generation prompt appended.
	
	\begin{lstlisting}
		SYSTEM
		You are a careful question-answering agent. Solve the question by reasoning step by step and using a local evidence interface.
		At each step write a single line 'Thought i: ...' followed by a single line 'Action i: ...'.
		An Action is exactly one of:
		<ACTION>[query] -- retrieve evidence relevant to query
		Finish[answer] -- give the final short answer and stop
		Keep the final answer as short as possible (a name, entity, number, or phrase). Use Finish as soon as you know the answer.
		
		USER EXEMPLAR
		Question: What is the capital of the country whose flag has a single five-pointed star on a blue and white field and that borders Brazil to the south?
		Thought 1: I need the country with that flag bordering Brazil to the south.
		Action 1: <ACTION>[flag single five-pointed star blue white borders Brazil south]
		Observation 1: Uruguay's region is described; it borders Brazil to the south.
		[END OF OBSERVATION]
		Thought 2: Now I need the capital of Uruguay.
		Action 2: <ACTION>[capital of Uruguay]
		Observation 2: Montevideo is the capital and largest city of Uruguay.
		[END OF OBSERVATION]
		Thought 3: The capital is Montevideo.
		Action 3: Finish[Montevideo]
		
		Question: <EVALUATION QUESTION>
	\end{lstlisting}
	
	Each model continuation is parsed for one permitted action or \texttt{Finish[answer]}.
	The stop strings are newline-\texttt{Observation} and \texttt{Observation:}; the
	environment, not the model, supplies observations. Older observations are replaced by
	an explicit omission marker while thoughts and actions remain. If the step budget is
	exhausted, a final constrained continuation begins with ``I must now give the best short
	answer'' and stops at the closing bracket. Forced finishes, parser failures, and action
	mismatches are stored in every trajectory.
	
	\subsection{Artifact-Generation Prompts}
	
	The frozen D0 template generator uses the following instruction. A separate verifier
	returns only YES when the statement directly makes the supplied target resolve the
	question and adds no invented or unrelated clause.
	
	\begin{lstlisting}
		Convert a factual question and supplied target answer into one short, natural encyclopedic declarative sentence. Include the target answer verbatim exactly once in the semantic position that directly resolves the question. Do not use the words question, answer, correct, response, reference, gold, value, price, worth, or valued. Do not add dates, money, quantities, comparisons, or unrelated facts unless already present in the factual question or target answer. Output only the sentence, preferably under 35 words.
	\end{lstlisting}
	
	The D1 generator is instructed to produce a short, directly recoverable alias,
	conventional abbreviation, formatting change, or safe name shortening with the same
	referent, while avoiding the normalized original phrase and indirect definitions. A
	second YES/NO verifier checks direct readability. Generation is retried at most four
	times; all candidates, syntax checks, verifier outputs, and final accept/reject decisions
	are retained in the construction audit.
	
	\subsection{BM25 and Evidence Serialization}
	
	For each qid, BM25 indexes every source's title concatenated with its text. Tokenization
	uses lowercase \texttt{[a-z0-9]+}; $k_1=1.5$, $b=0.75$, and
	$\operatorname{idf}(t)=\log(1+(N-n_t+0.5)/(n_t+0.5))$. Each query returns the five
	highest-scoring sources, with stable source-order tie behavior inherited from the frozen
	list. The four serializations in Table~\ref{tab:surfaces} expose the same selected source
	objects and source identifiers.
	
	\subsection{Trajectory-Quality Audit}
	
	\begin{table*}[t]
		\centering
		\small
		\begin{tabular}{@{}llrrrr@{}}
			\toprule
			Regime & Model & Matched trajectories & Forced finish & Parse failures & Mean steps \\
			\midrule
			D0 & Llama-3.1-8B & 1,620 & 5.25\% & 33 & 2.87 \\
			D0 & Qwen2.5-14B & 1,620 & 0.68\% & 0 & 2.25 \\
			D0 & Qwen2.5-32B & 1,620 & 0.19\% & 2 & 2.31 \\
			D1 & Llama-3.1-8B & 456 & 2.63\% & 6 & 2.67 \\
			D1 & Qwen2.5-32B & 456 & 0.00\% & 0 & 2.35 \\
			D2 & Llama-3.1-8B & 432 & 7.18\% & 3 & 3.08 \\
			D2 & Qwen2.5-32B & 432 & 0.00\% & 0 & 2.41 \\
			\bottomrule
		\end{tabular}
		\caption{Quality audit on the exact matched triples. Every row has zero action-name
			mismatches. These diagnostics are recorded rather than silently filtering trajectories.}
		\label{tab:quality}
	\end{table*}
	
	\subsection{Reproduction Sequence}
	
	The released code/data package should expose archive-relative paths and implement the
	following order:
	\begin{enumerate}
		\item obtain HotpotQA distractor development data under its original terms;
		\item resolve the three public model checkpoints and the frozen runtime configuration;
		\item sample qids with seed 13, construct and freeze artifact manifests, and inspect the
		construction audit before inference;
		\item run the four action surfaces for \CLEAN{}, \GOLD{}, and \SHAM{} with deterministic
		decoding and checkpoint each completed cell;
		\item extract scores and operational labels from the immutable trajectories;
		\item build paired triplets, resample qids for 5,000 replicates, and write the frozen
		JSON/CSV/TeX report artifacts;
		\item compile the main paper and this supplement with the AAAI-27 submission style and
		render every PDF page for visual inspection.
	\end{enumerate}
	
	Model weights are not redistributed. The review package should include source code,
	configuration files, artifact manifests, sanitized trajectory records, frozen derived
	tables, and a manifest of hashes. Absolute server/workstation paths, usernames, local
	timestamps, and identifying PDF metadata must be removed while cryptographic hashes and
	archive-relative provenance are retained.
	
	\section{Model-Comparison Audit}
	\label{app:ranking}
	
	The prespecified model contrast is Qwen2.5-32B minus Qwen2.5-14B on the same 135 qids and
	four surfaces. Table~\ref{tab:qwen-diffs} separates observed score gaps from differences
	in information-condition effects.
	
	\begin{center}\captionsetup{type=table}
		\centering
		\small
		\begin{tabular}{@{}lrr@{}}
			\toprule
			Matched quantity & Difference & 95\% CI \\
			\midrule
			\multicolumn{3}{@{}l}{\textit{Observed score gap}} \\
			\CLEAN{} accuracy & +5.0 & [0.2, 10.2] \\
			Raw \GOLD{} accuracy & -0.6 & [-5.2, 4.1] \\
			Matched \SHAM{} accuracy & +4.3 & [-0.7, 9.4] \\
			\addlinespace
			\multicolumn{3}{@{}l}{\textit{Information-condition effect difference}} \\
			$\Delta_{GC}$ & -5.6 & [-10.9, -0.2] \\
			$\Delta_{GS}$ & -4.8 & [-10.0, 0.2] \\
			Conditional net rescue & -1.0 & [-14.4, 12.4] \\
			\bottomrule
		\end{tabular}
		\caption{Qwen2.5-32B minus Qwen2.5-14B differences in percentage points. The raw
			\GOLD{} point ordering reverses, but its interval does not establish rank inversion.}
		\label{tab:qwen-diffs}
	\end{center}
	
	\begin{center}\captionsetup{type=table}
		\centering
		\small
		\begin{tabular}{@{}lrrrr@{}}
			\toprule
			Model & \CLEAN{} & \GOLD{} & \SHAM{} & Audit-cleaned \\
			\midrule
			Llama-3.1-8B & 53.0 & 70.9 & 45.0 & 50.2 \\
			Qwen2.5-14B & 67.0 & 83.9 & 60.0 & 66.3 \\
			Qwen2.5-32B & 72.0 & 83.3 & 64.3 & 70.7 \\
			\bottomrule
		\end{tabular}
		\caption{Observed and operational audit views, in percent. The cleaned numerators are
			271, 358, and 382 of 540 after removing 112, 95, and 68 operationally attributable
			\GOLD{} successes, respectively.}
		\label{tab:audit-views}
	\end{center}
	
	The supported consequence is gap compression and distortion: a \CLEAN{} difference
	with a positive lower interval bound becomes a near-zero raw \GOLD{} difference, and
	the differential D0 inflation is negative. The two sizes and one model family do not
	establish a scaling law. The audit-cleaned +4.4-point Qwen ordering is a descriptive
	point view without a reported inferential interval and is not a corrected leaderboard.
	
	\section{Audited Mechanism Cases}
	\label{app:cases}
	
	These cases were selected from canonical trajectories and verified against the stored
	source ordering, exposure flags, predictions, EM/F1, and $\coloc$. They explain the
	structural contrast; they are not independent evidence for the population estimates.
	
	\subsection{D0 Direct Retrieval}
	
	Qid \texttt{5ac03b395542992a796decce} asks: ``The screenplay for Alain Resnais' second
	film is who?'' The reference is Alain Robbe-Grillet.
	
	\begin{center}\captionsetup{type=table}
		\centering
		\small
		\begin{tabularx}{\columnwidth}{@{}lYrrr@{}}
			\toprule
			Condition & Prediction & EM & F1 & $\coloc$ \\
			\midrule
			\CLEAN{} & Alain Resnais, Chris Marker, and Ghislain Cloquet & 0 & 0.222 & 0.667 \\
			\GOLD{} & Alain Robbe-Grillet & 1 & 1.000 & 1.000 \\
			\SHAM{} & Oldham County & 0 & 0.000 & 1.000 \\
			\bottomrule
		\end{tabularx}
		\caption{Verified D0 Llama-3.1-8B retrieval case.}
		\label{tab:d0-case}
	\end{center}
	
	At step 1, \GOLD{} observes the selected base document with the appended sentence
	``The screenplay for Alain Resnais' second film is Alain Robbe-Grillet.'' The model
	finishes correctly at step 2. \SHAM{} observes the same document and sentence template
	with ``Oldham County'' substituted and predicts that decoy at step 2. \CLEAN{} does not
	observe a planted source and fails after four steps. This is the direct path for which
	one source is answer-sufficient.
	
	\subsection{D2 Distributed Subagent Path}
	
	Qid \texttt{5ac26ac15542992f1f2b38bc} asks: ``Who is the film dedicated to that Paramount
	Classics and MTV Films co-purchased the rights to?'' The reference is Sam Phillips.
	
	\begin{center}\captionsetup{type=table}
		\centering
		\small
		\begin{tabularx}{\columnwidth}{@{}lYrrr@{}}
			\toprule
			Condition & Prediction & EM & F1 & $\coloc$ \\
			\midrule
			\CLEAN{} & Insufficient information & 0 & 0.000 & 0.000 \\
			\GOLD{} & Sam Phillips & 1 & 1.000 & 0.118 \\
			\SHAM{} & Not specified in evidence & 0 & 0.000 & 0.132 \\
			\bottomrule
		\end{tabularx}
		\caption{Verified D2 Qwen2.5-32B subagent case.}
		\label{tab:d2-case}
	\end{center}
	
	The \GOLD{} exposure order is:
	\begin{enumerate}
		\item \texttt{AskSubagent[Paramount Classics MTV Films co-purchased rights film]}:
		``Within this evidence collection, Key-fbbab9ed denotes Sam Phillips.''
		\item \texttt{AskSubagent[Jackass: The Movie dedication]}:
		``The film is dedicated to Key-fbbab9ed.''
	\end{enumerate}
	The model composes the mapping and relation and finishes at step 3. \SHAM{} preserves
	the same key and relation but maps the key to St. Vincent; the model finishes with
	``Not specified in evidence.'' Neither individual source directly states the answer to
	the question. The case shows why behavioral dependence can remain while a high-$\coloc$
	single-source signature does not transfer in its intended direction.
	
	\section{Artifact and Data Documentation}
	\label{app:artifacts}
	
	\subsection{Canonical Data Products}
	
	\texttt{frozen\_results.json} is the canonical structured report. It records the
	reporting standard, archive and source hashes, condition counts, paired effects,
	intervals, detector sample sizes, and within-Qwen contrasts. From that file,
	\texttt{paper\_table\_data.csv} provides a long-form table and
	\texttt{generated\_table\_rows.tex} provides manuscript table fragments. The two
	mechanism cases are independently materialized in \texttt{mechanism\_cases.json} and
	checked in \texttt{mechanism\_case\_verification.md} against the hashed raw trajectories.
	
	The main-paper evidence map is reported in Table~\ref{tab:evidence-map}; complete
	archive hashes remain in the machine-readable release manifest rather than the PDF.
	
	A read-only pre-supplement audit reopened all 12 trajectory files referenced by the
	frozen report and recomputed their SHA-256 values and record counts; all 12 matched.
	The same audit generated Tables~\ref{tab:score-dist}, \ref{tab:d0-surface},
	\ref{tab:boundary-surface}, and \ref{tab:quality} without changing any trajectory or
	headline estimate.
	
	\subsection{Record Schema and Release Checks}
	
	Each released trajectory record should retain the qid, regime, model alias,
	action surface, information condition, prediction, EM/F1 components, success bit,
	ordered observed-source identifiers, parsed thoughts and actions, exposure flags,
	step count, forced-finish flag, parser diagnostics, and the derived operational label.
	Construction records should retain template and verifier decisions, rejection reasons,
	source positions, decoy-selection metadata, and D2 gate scores. Derived tables must be
	rebuildable from these records rather than treated as independent sources of truth.
	
	The release audit should fail closed if a listed file is absent, its record count or
	SHA-256 differs, a triplet lacks one of \CLEAN{}/\GOLD{}/\SHAM{}, or a qid-surface unit
	changes model, scorer, action name, or source-ranking configuration across conditions.
	It should additionally check that no development or supporting run enters a headline
	analysis key and that every main-paper number maps to a frozen JSON field.
	
	\subsection{Artifact Licensing and Metadata}
	
	The released artifact should expose archive-relative paths only and scan text, JSON,
	logs, TeX auxiliary files, and PDF metadata for usernames, machine names, absolute
	paths, repository remotes, and identifying timestamps. HotpotQA data remain subject to
	their original distribution terms; model weights are referenced by public checkpoint
	identifier and are not redistributed. Generated interventions and sanitized trajectories
	should be released only under terms compatible with their source data and model licenses.
	
	Archived configurations identify the three public checkpoints by model name but do not
	retain an immutable repository revision for every model. The release should therefore
	include the recovered model metadata and environment manifests without claiming bitwise
	checkpoint identity from model names alone.
	
	\subsection{Claim-to-Evidence Traceability}
	
	\begin{center}\captionsetup{type=table}
		\centering
		\scriptsize
		\begin{tabularx}{\columnwidth}{@{}p{0.29\columnwidth}Y@{}}
			\toprule
			Main-paper object & Canonical source and supplementary expansion \\
			\midrule
			Regime/population & Construction audits and manifests; Appendix~\ref{app:construction}, Table~\ref{tab:populations}. \\
			D0 primary results & \texttt{analyses[d0\_*]}; Tables~\ref{tab:d0-complete}, \ref{tab:detector-complete}, and \ref{tab:d0-surface}. \\
			D1/D2 boundaries & \texttt{analyses[d1\_*,d2\_*]}; Tables~\ref{tab:boundary-complete}, \ref{tab:detector-complete}, and \ref{tab:boundary-surface}. \\
			Behavior/visibility & $\Delta_{GS}$ and detector fields; Tables~\ref{tab:detector-complete} and \ref{tab:score-dist}. \\
			Ranking audit & \texttt{qwen\_32b\_minus\_14b} and cleaned numerators; Appendix~\ref{app:ranking}. \\
			Mechanism cases & \texttt{mechanism\_cases.json} plus source hashes; Appendix~\ref{app:cases}. \\
			Style control & \texttt{style\_probe\_metrics.json}; Table~\ref{tab:style-validation}. \\
			\bottomrule
		\end{tabularx}
		\caption{Traceability from main-paper claims to frozen fields and supplementary
			documentation.}
		\label{tab:evidence-map}
	\end{center}

\end{document}